# Language Models have higher Agreement than Humans in Historical Interpretation

Fabio Celli
*Research & Development*
Maggioli SpA
Santarcangelo di Romagna, Italy
fabio.celli@maggioli.it

Georgios Spathoulas
*Department of Information Security and Communication Technology*
Norwegian University of Science and Technology - NTNU
Gjøvik, Norway
georgios.spathoulas@ntnu.no

*Abstract*—This paper compares historical annotations by humans and LLMs. The findings reveal that both exhibit some cultural bias, but LLMs achieve a higher consensus on the interpretation of historical facts from short texts. While humans tend to disagree on the basis of their personal biases, LLMs disagree when they skip information or produce hallucinations. These findings have significant implications for digital humanities, enabling large-scale annotation and quantitative analysis of historical data. This offers new educational opportunities to explore historical data in an interactive way, but it also poses threats to collective memory, such as narrative homogenization, manipulation, or even selective censorship.

## I. Introduction

Interpreting historical dynamics is a very difficult task, where subjectivity, prejudice, and Eurocentrism may introduce bias. Research reveals that historical accounts can be skewed by various influences, such as political motives, cultural beliefs, limited information, and the historian's own viewpoints. In general, bias can be corrected by reaching a consensus or a synthesis between different points of view, through peer review or other inclusive historical methods [1], but cultural biases can be difficult to recognize and overcome [16]. Understanding the mechanisms that influence historical interpretation is even more important in the age of post-truth, as there is an increasing amount of fake news and misinformation in the media that leads to increased polarization of opinions [15]. In that context, Large Language Models (LLMs) may also contain biases due to fake news and misinformation present in digital data they are trained on.

In this paper, we address the issue of historical data interpretation by comparing humans and LLMs in the annotation of historical cycles labels (i.e.: growth, apex, decline, crisis) from short descriptions of historical events. We contribute to current research on history interpretation with a quantitative experimental design. In particular, our aim is to answer the following Research Questions:

- (RQ1) Do LLMs exhibit historical or cultural biases?

This research was supported by the European Commission, grant 101120657: European Lighthouse to Manifest Trustworthy and Green AI—ENFIELD.

- (RQ2) Can different LLMs reach a higher consensus than humans on historical interpretation?
- (RQ3) What types of errors do humans and LLMs make when interpreting historical data?

The paper is structured as follows: in Section II we provide some background with theoretical frameworks and data. In Section III we describe the experimental design, with annotation guidelines, prompts, and the results of the experiments. Finally, in Section IV, we draw our conclusion.

## II. Related Work

There are many approaches to the study of history, but in general terms they can be grouped into two broad and fuzzy macro-categories. On the one hand, there is cultural/social history that includes authors like Foucault [8] and Harari [12]; on the other hand there is economic history, including authors like Marx and Piketty [18]. Cultural/social history examines the development and evolution of cultures through time, including artifacts, institutions, ideologies, and revolutions; economic history investigates economic structures and processes throughout history, including production, trade, finance, and technological change. In our experiment design, we selected one set of labels for cultural/social history, and one for economic history, in order to evaluate the annotation of LLMs and humans with quantitative methods on both the macro-categories. Previous attempts to compare human and artificial intelligence in historical interpretations contributed to define the experimental settings (evaluating the response of LLMs against human-annotated data) [3], [13] and found that agreement between LLMs depend on the model size [5].

### A. Historical Labels

On the side of cultural/social history, we selected the set of labels from structural-demographic theory (SDT), a framework to understand the dynamics of social and political instability in complex societies [10]. SDT has been applied to a wide range of historical cases, including the French Revolution, the American Civil War [20], the fall of the Qing Dynasty [17], as well as the instability in the US in recent years [21]. SDT posits that social unrest arises from the interplay of 3 actors: the State, with its capacity to maintain order and provide public goods; the elites with their wealth, knowledge, power, and internal divisions; and finally the general population with its demographics and cultures. These actors influence each other in a cyclical manner through feedback loops. For example, a growing population can strain state resources, leading to increased taxation and elite competition, which can further exacerbate social unrest. Interestingly, unrest is considered a quantitative feature, measurable by the number of rebellions, protests, and revolutions events in a society. Using scientific methods and mathematical modeling, it is possible to identify regularities in the rise and fall of polities over time. Societies typically last one or more secular cycles of variable length [22]. Secular cycles have different phases, gradually increasing social and political instability. Hoyer [14] proposed five phases for the secular cycle, reported in Table I:

TABLE I
PHASES OF THE SECULAR CYCLE (SDT LABELS)

| Label | Phase | Description | Examples |
|---|---|---|---|
| 0 | Crisis, Collapse, or Recovery | Elite-led reforms or overthrow, new social equilibrium. | France in 1790s, China in 1950s |
| 1 | Growth | Dynamic period with strong culture, rapid growth, expanding state control. | Post-war Italy (1950s) |
| 2 | Population Impoverishment | Demographic growth, economic slowdown, wage depression, elite enrichment, social unrest. | United States in 1890s and 1970s |
| 3 | Elite Overproduction | Increased access to elite ranks, overloaded social mobility, declining problem-solving capacity. | USSR in 1950s |
| 4 | State Stress | Weakened governance, elite fragmentation, eroded cooperation, potential violence. | Germany in the 1920s |

On the side of economic history, we selected the Big Cycle model. Studying the rise and fall of empires in the last 500 years from an economic point of view, Dalio [7] identifies three phases reported in Table II:

Big Cycles are driven by the interplay of debt, internal and external instability. As economies expand, credit and debt levels increase, and the distribution of wealth becomes unequal, so productivity decreases with negative impact on economic stability and social cohesion, triggering internal and external instability until the economic cycle restarts. The rise phase roughly corresponds to the growth phase, apex corresponds to the population impoverishment and elite overproduction phases, while decline corresponds to the State stress and crisis phases.

In our experimental design, we instruct LLMs and humans to annotate these labels from data describing historical periods, reported in the next Section.

TABLE II
PHASES OF THE BIG CYCLE.

| Label | Phase | Description | Examples |
|---|---|---|---|
| 0 | Rise | A strong society leverages over its competitive advantage and creates peace and prosperity | Dutch golden age around 1600s |
| 1 | Apex | A society must expend increasing resources with reduced returns to maintain its prosperity | US around the 2000s |
| 2 | Decline | A society can no longer meet the costs of maintaining its prosperity and other societies challenge its power | British Empire in the 1940s |

*B. Data*

There are historical datasets annotated with events [19] or with metadata about cultural phenomena [2], and it is even possible to produce synthetic data with annotations [6], but there is one annotated with SDT labels: the Chronos dataset [4]. The Chronos dataset includes short texts describing polities decade-by-decade from prehistory to 2010 CE in 18 sampling points, including Egypt, Iraq, China, Japan, Italy, Greece, India, Indonesia, Mexico, Hawaii and many others. To verify whether the labeling was consistent, the authors validated the annotation with three people, who independently labeled a sample of 93 examples from the data. They report that the initial agreement was low (Fleiss' $k$ 0.206) because a single disagreement has an exponential impact on the rest of the sequence, but after a training session and the use of a standard pattern to start with (the sequence of secular cycle labels 1,1,2,2,3,3,4,4,4,0), the agreement between humans raised to Fleiss' $k$ 0.455. In the same task, small LLMs have very poor agreement (Fleiss' $k$ 0.069), but very large models can reach incredibly high agreement (Fleiss' $k$ 0.808). We take these inter-annotator agreement values as reference for our experiments.

We sampled about 100 decades from different polities in different continents and historical periods: contemporary Europe from 1950s to 2010s, People's Republic of China from 1950s to 2010s, Mexican Republic from 1820s to 2010s, Later Jin Dynasty from 1130s to 1230s, Mongol Empire from 1240s to 1270s, Yuan Dynasty from 1280s to 1360s, Roman Kingdom from 750s BC to 500s BC and Roman Republic from 490s BC to 280s BC. Each data point contains a timestamp, a polity identifier, and a short description of about 300 characters. Examples follows:

> Decade: 1170s, polity: CnLrJin, description: Emperor Shizong (r. 1161-1189) confiscated unused land from Jurchen landowners and redistributed to the Jurchen farmers but they preferred to lease the work to Chinese farmers and engage in drinking instead. in 1175 paper factory in Hangzhou employed more than a thousand Chinese workers. Integration problems (language and customs).

Decade: -600s, polity: ItRomRg, description: Lucius Tarquinius Priscus waged wars against the Sabines and Etruscans, doubling the size of Rome and bringing great treasures to the city. To accommodate the influx of population, the Aventine and Caelian hills were populated.

Decade: 1960s, polity: CnPrcr*, description: from 1966 to 1976 the turbulent period of political and social chaos within China known as the Cultural Revolution (cancel capitalist and traditional culture with violence) led to greater economic and educational decline with millions being purged or subjected to either persecution or politicide. 10 millions of students were sent in the country as farmers. War with India for borders.

This latter example was selected in order to answer to RQ1 (Do LLMs present historical cultural bias?), as the interpretation of the 1960s in China makes evident a Eurocentric bias [4]. In particular, SDT (and Chinese historians) interpret this decade as a "growth" phase, because there was a new fresh culture that created social cohesion, but the Eurocentric interpretation is "crisis", because there was widespread violence. Hence, humans or LLMs that interpret this example as a "crisis" or "decline" phase show a historical cultural bias. We will see the results in the next Section.

## III. EXPERIMENTS

The experimental setting is simple: given the same prompt/guidelines to 3 LLMs and 3 humans, we asked them to annotate the selected examples, providing the label and an explanation of the decision for the two tasks: annotation of Secular Cycle (5 labels task) and Big Cycle (3 labels task). Previous literature reports that LMMs with fewer than 100 billion parameters are prone to hallucinate in the 5 labels task, while LLMs with more than 400 billion parameters can reach very high agreement [5]. We randomly selected three LLMs having more than 100 billion parameters, for instance : Claude 3.5 Sonnet (175 billion parameters), Gemini 1.5 Flash (number of parameters unknown, but likely more than 100 billions) and GPT-4 (1.8 trillion parameters). Figure 1 illustrates the workflow.

Starting from the Chronos dataset, we extracted 60 examples of historical decades descriptions per SDT label (reported in Table I) and summarized them with prompt 0, reported in Figure 2.

In this way, we used knowledge generation prompting [9] to extract cues associated with each label, and we inserted them in prompts 1 and 2. Prompts 1 and 2 are the guidelines for the annotation of secular cycles and big cycle phases used both by LLMs and humans. They are reported in Figures 3 and 4 respectively.

Two trials were conducted, one with the 5 secular cycle labels and one with the 3 big cycle labels, each with different human annotators. These annotators were students aged 20-25,

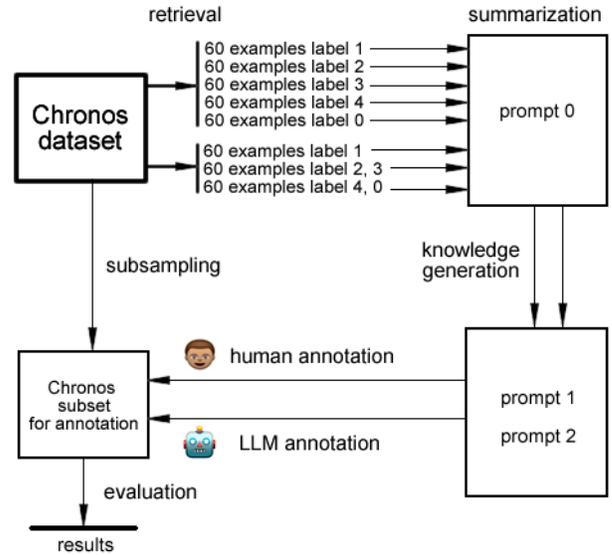

Fig. 1. Schema of the method adopted.

---
PROMPT 0
summarize the following phases into prompt instructions optimized to classify the phases reducing potential overlaps

---

Fig. 2. Prompt for the summarization of cues related to labels.

with no expertise in history. Two different triplets of human annotators were employed in the two trials, while the same LLMs were used for both tasks. We evaluated performance using two metrics: Fleiss' Kappa, which measures overall agreement, and the average correlation coefficient, which quantifies the similarity in trends across the annotated cycles. The evaluation was performed on a subset of 106 examples from the Chronos dataset described in Section II-B, the same subset used in previous literature [4].

In addition to the knowledge generation prompting technique used for the cues from labels, we used role-play prompting [11] ("Act as an expert historian"), and few-shot prompting ("Here is an example of the desired output: ...").

First, we checked the responses of humans and LLMs to the 1960s in China to answer RQ1, whose results are reported in Table III.

TABLE III
ANALYSIS OF HISTORICAL CULTURAL BIAS.

| settings | Gemini1.5f | Claude3.5s | GPT4 |
|---|---|---|---|
| 1960s CnPrcr* 5 labels | 0 (crisis) | 1 (growth) | 1 (growth) |
| 1960s CnPrcr* 3 labels | 2 (decline) | 0 (rise) | 0 (rise) |
| settings | annot.1 | annot.2 | annot.3 |
| 1960s CnPrcr* 5 labels | 0 (crisis) | 0 (crisis) | 0 (crisis) |
| 1960s CnPrcr* 3 labels | 2 (decline) | 2 (decline) | 2 (decline) |

Gemini 1.5 flash systematically showed historical cultural bias, annotating this phase as "crisis" and as "decline", respectively, in both the label schemas. GPT 4 and Claude 3.5 sonnet instead annotated this phase as "growth" and "rise"

```
PROMPT 1
Act as an expert historian and consider the Structural Demographic Theory (SDT).
Given a set of description of historical decades for different polities, label each
description with one of the following secular cycle phases, and provide a short
explanation of the choice:
0=crisis (in this phase may happen societal collapse patterns, power transitions,
conflicts, administrative changes, economic patterns, social structure changes, and
external influences. Look for signs of civil wars, military coups, environmental
factors, population movements, reform of tax systems, trade network disruptions, class
conflicts, and foreign invasions.
1=growth (a society recovers from a crisis finding a new refresh culture that creates
social cohesion. to recognize this phase examine the power structure patterns,
legitimacy of rule, social organization, economic patterns, cultural elements, military
aspects, and social changes. Look for the presence of elite classes, religious
legitimation of power, centralized administrative systems, trade networks, cultural
practices, territorial expansion, and population movements.),
2=population immiseration (growth slows and inequalities begin to emerge. to
recognize this phase evaluate the power dynamics, economic patterns, military aspects,
cultural/religious elements, administrative features, and infrastructure development.
Look for succession struggles, trade route development, territorial conquests, religious
tolerance, bureaucratic reforms, and construction projects.),
3=elite overproduction (the number elite aspirants rises, the social lift mechanisms
deteriorate. to recognize this phase assess power dynamics, governance, economic
patterns, social structures, crisis patterns, cultural and technological development, and
common catalysts for change. Look for power struggles, trade system developments,
elite class formation, military defeats, religious developments, and military conflicts).
4=state stress (elites struggle to institutionalize their advantages. to recognize this
phase review political instability, power struggles, economic challenges, military
conflicts, administrative changes, and social/religious tensions. Look for succession
disputes, financial crises, territorial expansions, bureaucratic reforms, and religious
conflicts.)
A cycle cannot turn back and cannot skip phases. So if in 1940 there is a phase 0, in
1950 there should be a phase 1, in 1960 there can be a phase 1 or phase 2. If in 1960
there is a phase 2, in 1970 there can be a phase 2 or phase 3, not a phase 4. If in
1970 there is a phase 3, in 1980 there can be a phase 3 or 4, and if in 2000 there is
phase 4, in 2010 there can be a phase 0 or another phase 4. The decade after phase 0
the cycle restarts from phase 1.
This is an example of the desired output:

decade,polityid,label,explanation
1940,UsUsar*,0,"(crisis) - war against Nazi Germany. Victory in WW2"
1950,UsUsar*,1,"(growth) - Expansion to Europe with Marshall Plan, economic
growth, civil right movements"
1960,UsUsar*,1,"(growth) - Continued strong growth, space race, Cuban missile
crisis"
1970,UsUsar*,2,"(immiseration) - Oil crisis, social tensions, inflation"
1980,UsUsar*,2,"(immiseration) - Recession, welfare retrenchment"
1990,UsUsar*,3,"(elite overproduction) - Invasion of Kuwait, power consolidation"
2000,UsUsar*,3,"(elite overproduction) - Terrorism, economic bubbles, financial crisis"
2010,UsUsar*,4,"(state stress) - Debt crisis, misinformation, social unrest"
```

Fig. 3. Prompt for the annotation of historical data with the 5 labels of the secular cycle.

```
PROMPT 2
Act as an expert historian and consider the economic cycles. Given a set of description
of historical decades for different polities, label each description with one of the
following big cycle phases, and provide a short explanation of the choice:
0=rise (a society recovers from a crisis finding a new refresh culture that creates social
cohesion. to recognize this phase examine the power structure patterns, legitimacy
of rule, social organization, economic patterns, cultural elements, military aspects,
and social changes. Look for the presence of elite classes, religious legitimation of
power, centralized administrative systems, trade networks, cultural practices, territorial
expansion, and population movements.),
1=apex (growth slows and inequalities begin to emerge. the number elite aspirants
rises, the social lift mechanisms deteriorate.. to recognize this phase evaluate the
power dynamics, economic patterns, military aspects, cultural/religious elements,
administrative features, and infrastructure development. Look for succession struggles,
trade route development, territorial conquests, religious tolerance, bureaucratic reforms,
and construction projects. Also assess power dynamics, governance, economic patterns,
social structures, crisis patterns, cultural and technological development, and common
catalysts for change. Look for power struggles, trade system developments, elite class
formation, military defeats, religious developments, and military conflicts).
2=decline (in this phase may happen societal collapse patterns, power transitions,
conflicts, administrative changes, economic patterns, social structure changes, and
external influences. Look for signs of civil wars, military coups, environmental
factors, population movements, reform of tax systems, trade network disruptions,
class conflicts, and foreign invasions. To recognize this phase also review political
instability, power struggles, economic challenges, military conflicts, administrative
changes, and social/religious tensions. Look for succession disputes, financial crises,
territorial expansions, bureaucratic reforms, and religious conflicts.)
A cycle cannot turn back and cannot skip phases. So if in 1940 there is a phase 0, in
1950 there should be a phase 1, in 1960 there can be a phase 1 or phase 2. If in 1960
there is a phase 2, in 1970 there can be a phase 2 or phase 0. A phase 0 cannot come
after a phase 1.
Here is an example of the desired output:

decade,polityid,label,explanation
1940,UsUsar*,2,"(decline) - war against Nazi Germany. Victory in WW2"
1950,UsUsar*,0,"(growth) - Expansion to Europe with Marshall Plan, economic
growth"
1960,UsUsar*,0,"(growth) - Continued strong growth, space race, Cuban missile
crisis"
1970,UsUsar*,1,"(apex) - Oil crisis, social tensions, inflation"
1980,UsUsar*,1,"(apex) - Recession, welfare retrenchment"
1990,UsUsar*,1,"(apex) - Invasion of Kuwait, power consolidation"
2000,UsUsar*,1,"(apex)- Terrorism, economic bubbles, financial crisis"
2010,UsUsar*,2,"(decline) - Debt crisis, misinformation, social unrest"
```

Fig. 4. Prompt for the annotation of historical data with the 3 labels of the Big Cycle.

TABLE IV
COMPARISON OF LLMS AND HUMANS IN HISTORICAL ANNOTATION.
THE BEST RESULTS ARE MARKED IN BOLD.

| settings | Fleiss $K$ | avg $\rho$ |
|---|---|---|
| 3 LLMs, 5 lables, 106 examples | **0.330** | **0.344** |
| 3 LLMs, 3 lables, 106 examples | 0.228 | **0.419** |

| settings | Fleiss $K$ | avg $\rho$ |
|---|---|---|
| 3 humans, 5 lables, 106 examples | 0.091 | 0.008 |
| 3 humans, 3 lables, 106 examples | **0.253** | 0.327 |

with the secular cycle and big cycle respectively, showing no cultural bias. All human annotators were Italian students with different backgrounds, and, crucially, they showed a consistent historical interpretation bias.

In order to answer RQ2 (Can different LLMs reach a higher consensus than humans on historical interpretation?), we compared the inter-annotator agreement and averaged correlation coefficient among 3 humans and 3 LLMs. The results, reported in Table IV, show that LLMs demonstrate significantly better agreement on historical annotations than humans when interpreting the 5 labels of the secular cycle.

However, the picture is different for the 3 labels of the big cycle. In this case, human and machine agreement is comparable, with humans achieving a slightly higher level (0.253). However, even here, LLMs exhibit a higher average correlation coefficient. This suggests that machine-generated annotations tend to identify longer uninterrupted periods, whereas human annotations reflect more fragmented agreement across decades.

The agreement between human annotators in the 5 labels task obtained here (0.091) is much lower than Fleiss' K=0.206 reported in previous literature [4]. This suggests that there is more variability than previously reported in the annotation of historical labels, depending on the background culture of the annotators and the clarity of the guidelines.

Given this variability, we decided to replicate the experiment with our annotators using the conventional sequence schema reported in literature that obtained K=0.455 with 3 human annotators on 93 examples. We added this sentence to the prompt and run the experiment also with the LLMs: "Initially assume the sequence of labels follows the one of the standard secular cycle model: 1,1,2,2,3,3,4,4,4,0 and then evaluate whether to keep or change the labels in each decade. It is possible to have longer or shorter cycles". The results, reported in Table V, reveal that LLMs also beat human annotators with this interpretive schema, although the agreement score decreases. It is interesting to note that using the conventional sequence schema, the Fleiss' K increased for humans by ≈0.25 points, and this is consistent with the results reported in the literature. Again, this confirms the variability of human annotation, that may depend on personal biases, like opinions and political views. Potentially, using like-minded human annotators it is possible to reach moderate agreement, but when humans have different views, the agreement score drops to slight or no agreement.

TABLE V
REPLICATION OF EXPERIMENT WITH SEQUENCE SCHEMA.
THE BEST RESULTS ARE MARKED IN BOLD.

| settings | Fleiss $K$ | avg $\rho$ |
| --- | --- | --- |
| 3 humans, 5 lables, 93 examples | 0.258 | **0.405** |
| 3 LLMs, 5 lables, 93 examples | **0.274** | 0.366 |

Anyway, the effect of the conventional sequence schema on human annotation is to dramatically increase the average correlation coefficient, meaning that there are longer, uninterrupted decades annotated in the same way. The use of this schema with LLMs improves just a little bit the average correlation, at the cost of producing more sequence errors, thus decreasing Fleiss' K.

We ran another experiment to answer RQ3 (What types of errors do humans and LLMs make when interpreting historical data?). First of all, we counted the errors due to constraint violations (on the results of the original experiment with 106 examples). Prompt 1 does not allow multiple crisis phases, and both prompts do not allow for phase skipping and backward phases. The results, reported in Table VI, show that LLMs are more prone to skip phases, while humans did not respect the constraint of the two adjacent crisis phases.

TABLE VI
ANALYSIS OF ANNOTATION ERRORS.

| settings | repeated crises | skipping | backward phase |
| --- | --- | --- | --- |
| llms, 5 lables | 3 | 7 | 2 |
| llms, 3 lables | - | 0 | 3 |

| settings | repeated crises | skipping | backward phase |
| --- | --- | --- | --- |
| humans, 5 lables | 11 | 3 | 2 |
| humans, 3 lables | - | 0 | 2 |

In particular, Gemini tends to skip phases in the 5 labels task, but not in the 3 labels task. GPT tends to generate backward phases both with 5 and 3 labels. Claude generates fewer errors than other models. It is interesting to note that both humans and machines do not skip phases in the 3 labels task.

To understand what humans and machines found most difficult to annotate, we extracted the explanations where the 3 annotators disagreed, and plotted word clouds, reported in Figure 5.

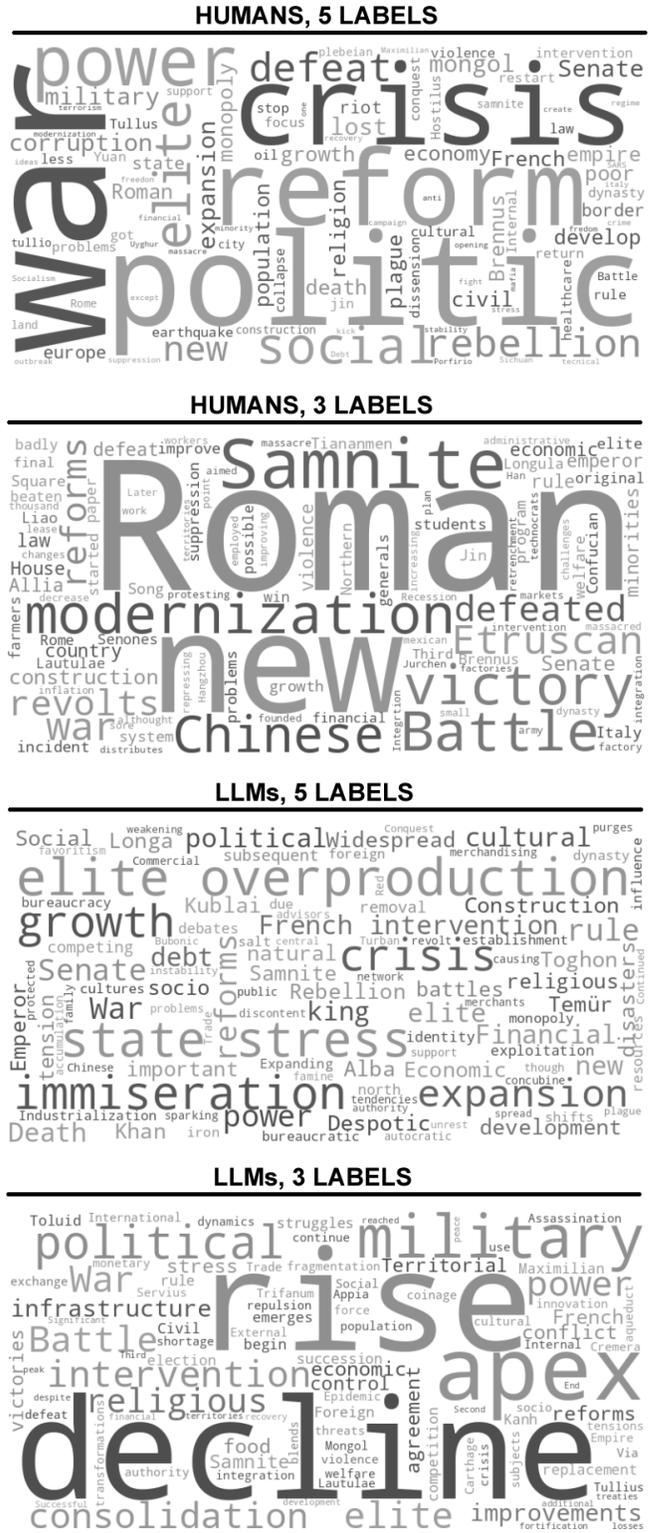

Fig. 5. Wordclouds of the explanations on examples where all 3 annotators were in disagreement.

These charts reveal that "war", "crisis", "reform" and "politics" are the most frequent words of the 5 labels task appearing in contexts where human annotators disagree, while "Roman" and "new" are proxies of difficult contexts in the 3 labels task. An interview with human annotators revealed that the application of economic cycles to roman history was particularly challenging because there was a lot of information about wars and few about the economy. LLMs always reported the label of choice in the explanation ("growth", "crisis", "immiseration", "rise", "decline" and so on). The wordclouds reveal that there are no single specific words associated to contexts of disagreement in the 5 labels task, while the "rise" and "decline" labels are the most difficult to annotate for machines in the 3 labels task.

## IV. CONCLUSION

The results presented in this paper demonstrate at least three important points:

1) Gemini 1.5 flash shows a historical cultural bias. Crucially, all human annotators showed the same kind of Eurocentric cultural bias without exception.
2) LLMs can reach a higher consensus than humans in historical interpretation, with or without the use of conventional sequence schemas. The best results are obtained with the use of LLMs, without using any sequence schema in the prompt. Moreover, LLMs are better than humans on the side of cultural/social history, the task that proved to be most difficult for human annotators because of the complexity of interweaving facts.
3) Both humans and LLMs make errors. While humans tend to put their own view in historical interpretation, potentially decreasing agreement with other annotators, LLMs tend to skip phases. Hallucinations are always possible, especially with small models, but we do not have any in our experiment since we used models larger than 100 billion parameters.

These findings have significant implications for digital humanities. While historians' expertise remains crucial for public history discussions, AI now provides powerful tools for annotating large historical datasets. This opens doors for more extensive, data-driven historical analysis, helping researchers reduce personal bias. LLMs and big data can revolutionize digital humanities by creating new educational experiences, like interactive history learning, exploring alternative historical scenarios, and fostering critical thinking about bias.

However, we must be aware of the risks. LLMs trained on biased or inaccurate data can spread misinformation and homogenize narratives. To address this, digital humanities and computational history need to collaborate with trustworthy AI experts. We should focus on making AI annotations transparent, using diverse historical datasets, and comparing outputs from different models. AI offers immense potential, but careful implementation is essential to maximize its benefits and minimize its risks.